\documentclass[10pt,twocolumn,letterpaper]{article}

\usepackage{iccv}
\usepackage{times}
\usepackage{epsfig}
\usepackage{graphicx}
\usepackage{amsmath}
\usepackage{amssymb}
\usepackage{algorithm}
\usepackage{algorithmicx}
\usepackage{algcompatible}
\usepackage{xcolor}
\usepackage{subcaption}

\usepackage[sort&compress,square,comma,numbers]{natbib}
\makeatletter
\renewcommand\bibsection%
{
  \section*{\refname
    \@mkboth{\MakeUppercase{\refname}}{\MakeUppercase{\refname}}}
}
\makeatother

\usepackage{color}
\usepackage{url}
\usepackage[pagebackref=true,breaklinks=true,letterpaper=true,colorlinks,bookmarks=false]{hyperref}
\hypersetup{colorlinks=true, linkcolor=red}

\iccvfinalcopy % *** Uncomment this line for the final submission

\def\iccvPaperID{6083} % *** Enter the ICCV Paper ID here

\ificcvfinal\fi

\begin{document}

%%%%%%%%% TITLE
\title{Imitation Learning for Human Pose Prediction}

\author{Borui Wang, Ehsan Adeli, Hsu-kuang Chiu, De-An Huang, Juan Carlos Niebles\\
Stanford University\\
\texttt{\small \{wbr, eadeli, hkchiu, dahuang, jniebles\}@cs.stanford.edu}
}

\maketitle

%%%%%%%%% ABSTRACT
\begin{abstract}
Modeling and prediction of human motion dynamics has long been a challenging problem in computer vision, and most existing methods rely on the end-to-end supervised training of various architectures of recurrent neural networks. Inspired by the recent success of deep reinforcement learning methods, in this paper we propose a new reinforcement learning formulation for the problem of human pose prediction, and develop an imitation learning algorithm for predicting future poses under this formulation through a combination of behavioral cloning and generative adversarial imitation learning. Our experiments show that our proposed method outperforms all existing state-of-the-art baseline models by large margins on the task of human pose prediction in both short-term predictions and long-term predictions, while also enjoying huge advantage in training speed.
\end{abstract}

%%%%%%%%% BODY TEXT
\section{Introduction}

Modeling the dynamics of human motion and predicting human poses is an important and challenging problem in computer vision that has many useful applications in robotics, computer graphics, healthcare, public safety, etc. \cite{Gupta14, Koppula13, Kovar02, Troje02, Urtasun06}. Previous work on this subject has mainly been focusing on designing different architectures of recurrent neural networks (RNNs) to model human motion dynamics and adopted a pure supervised-learning approach to train recurrent neural networks to predict future human poses in a sequence \cite{fragkiadaki2015recurrent, jain2016structural, martinez2017human, Chiu18}. 

These previous methods based on supervised training of RNN architectures face two main challenges: (1) due to the purely supervised nature of the training methods, the learned RNNs usually do not generalize well to unseen domains of the human motion space, which are very likely to appear during test time; (2) since the RNNs are required to generate the whole human pose prediction sequence all together, it is very difficult to keep a good balance between short-term and long-term prediction accuracies. 

\begin{figure}
\centerline{\includegraphics[width=\columnwidth]{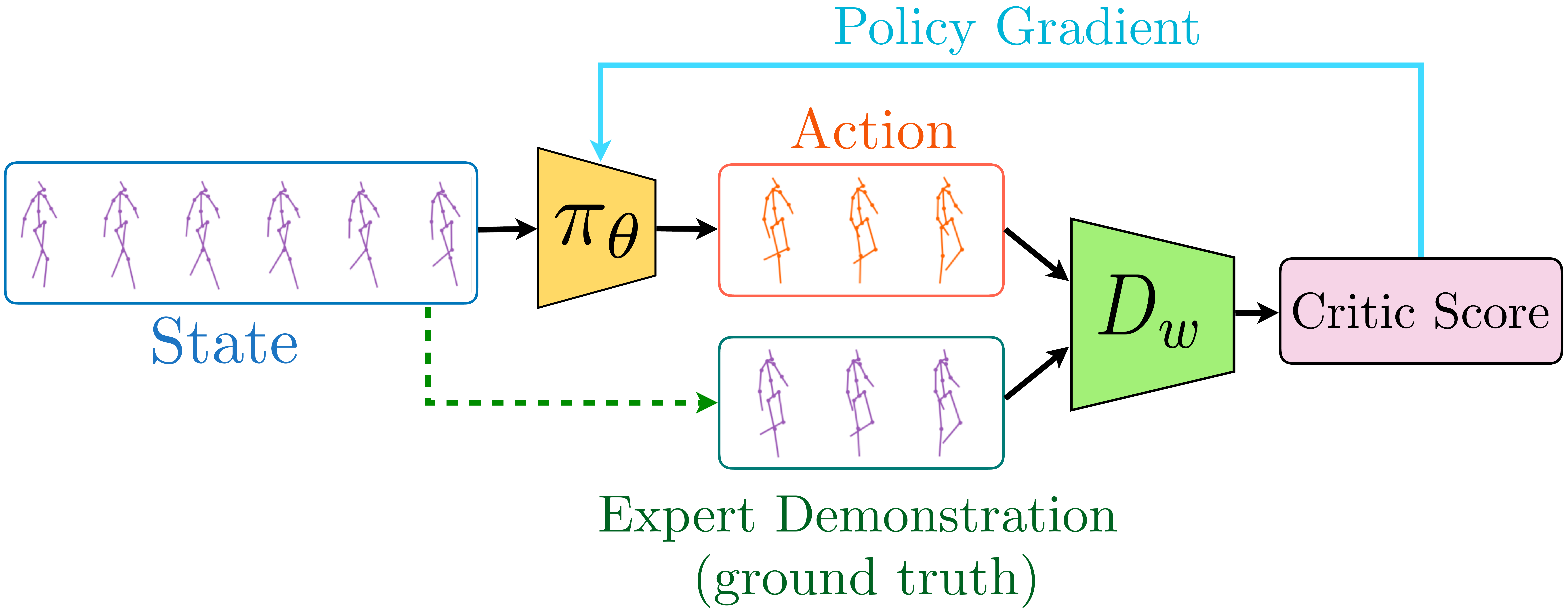}}
\caption{At the core of our imitation learning approach to human pose prediction is a Generative Adversarial Imitation Learning (GAIL) \cite{Ho16} process. With the critic function $D_w$ and the policy generator $\pi_{\theta}$, we alternate between updating $D_w$ ($\mathbf{D}$ step) through comparing generated windows of predicted poses with ground truth, and updating $\pi_{\theta}$ ($\mathbf{G}$ step) through policy gradient over critic scores from $D_w$.}\label{fig:1}
\vspace{-10pt}
\end{figure}
%\vspace{-10pt}

Recently, we have witnessed great success in the development of deep reinforcement learning algorithms, which have achieved significantly enhanced state-of-the-art performance on many tasks in the areas of game, robotics and control \cite{Mnih13, Schulman15, Schulman17}. These recent deep reinforcement learning algorithms, including Generative Adversarial Imitation Learning (GAIL) \cite{Ho16} and Deep Deterministic Policy Gradient \cite{Lillicrap16}, enjoy the advantages of generalizing well over unseen terrains and maintaining strong sequential correlation among local decisions across time. Therefore, in order to overcome the above limitations of the previous methods for human pose prediction, a natural question to ask is whether we could build a bridge between reinforcement learning and sequential modeling, such that we can harness the great power of deep reinforcement learning algorithms to help us better model human motion dynamics and predict human poses from sequential observations. In this paper, we propose a new modeling framework for human motion dynamics that transforms the task of predicting human poses into a reinforcement learning problem. The environment of this reinforcement learning problem cannot provide feedback signals as the learning agent interacts with it. All we have during the learning process is a training dataset consisting of trajectories of human poses recorded from real human motion. Therefore, we adopt an imitation learning \cite{Ross11, Zeng17} approach and use the training dataset as expert demonstrations for our prediction agent to imitate.

This imitation learning problem of predicting human poses has several key difficulties: (1) the action space is continuous-valued and very high-dimensional; (2) the state space is heterogeneous and encompasses very long sequences of historical observations with different lengths; (3) the expert demonstrations are performed by different human subjects and thus exhibit large variance across the underlying expert policies. To tackle this challenging imitation learning task, we extend the Generative Adversarial Imitation Learning (see Figure \ref{fig:1}) framework with sequence-to-sequence architectures \cite{Sutskever14} and Deep Deterministic Policy Gradient \cite{Lillicrap16} methods to train our prediction agent to make accurate predictions of human poses. We also use the efficient behavioral cloning \cite{Bain95} algorithm as pre-training to expedite the training process.

We evaluate the performance of our proposed imitation learning algorithm on the popular Human 3.6M dataset \cite{Ionescu14}. Our experiments demonstrate that our proposed algorithm outperforms all the previous methods for human pose prediction by large margins on both short-term and long-term predictions and sets the new state-of-the-art performance results on the Human 3.6M dataset. The experiments also show that our algorithm has huge advantage in speed and can be trained much more efficiently compared to previous algorithms. 

To summarize, the main contributions of our work are: (1) We propose a new reinforcement learning formulation for the problem of human pose prediction that supports more accurate predictions over both short-term and long-term horizons; (2) We develop an imitation learning algorithm for human pose prediction based on this reinforcement learning formulation through a combination of behavioral cloning and generative adversarial imitation learning. Our algorithm combines the advantages from both learning frameworks and achieves a good balance between sample efficiency and policy generalizability; (3) We run extensive experiments on the challenging Human 3.6M dataset to evaluate the performance of our proposed method, and show that it outperforms all existing state-of-the-art baseline models by significant margins.

\begin{figure}
\centerline{\includegraphics[width=\columnwidth]{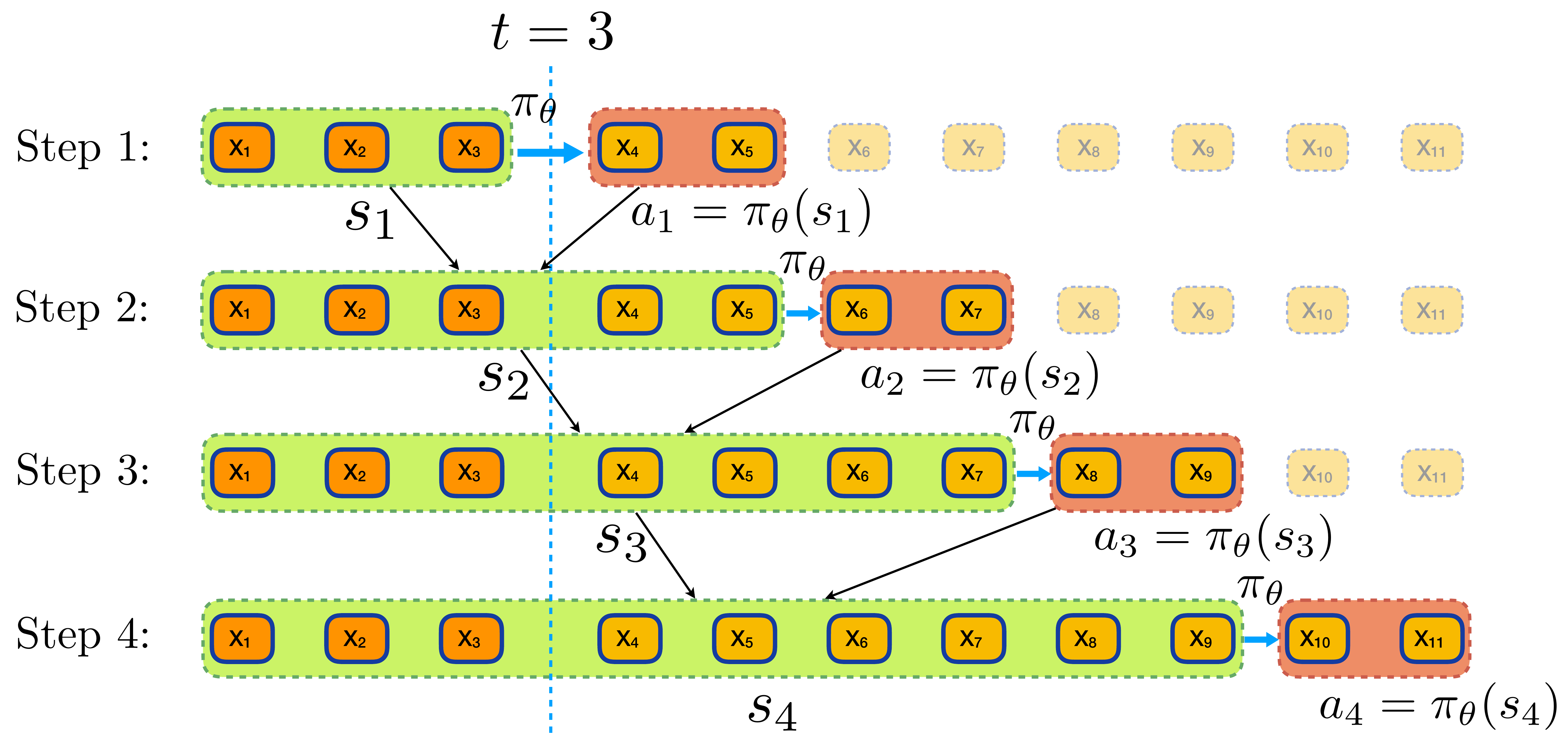}}
\caption{Illustration of progressive prediction and our reinforcement learning formulation of human pose prediction using an example pose prediction task with length of past observations $t = 3$,\, length of total future predictions $l = 8$  and\, size of step-wise prediction windows $m = 2$.} \label{fig:fig2}
\vspace{-10pt}
\end{figure}

\section{Related Work} 
\noindent\textbf{Human Motion Prediction:} 
Most of the previous work on video prediction predict future video sequences by reconstructing frames at the pixel level \cite{vondrick2016generating,mahjourian2017geometry}, and predict dense trajectories \cite{walker2015dense}, semantic labels \cite{luc2017predicting,walker2016uncertain}, or activity labels \cite{walker2017pose,soomro2018online,arzani2017structured,liang2019peeking} in the future. However, human motion dynamics is better captured by detailed joint locations (\ie, pose)  \cite{rogez2008randomized,sandriluka2010monocular,mehta2017vnect}, and is often modeled by either \textit{state transition models} \cite{wang2008gaussian,wu2014leveraging} or \textit{recurrent neural networks} \cite{ghosh2017learning,martinez2017human,jain2016structural}. 

Recently, several works focused on forecasting human poses in videos \cite{jain2016structural,martinez2017human,Chiu18,chao2017forecasting,walker2017pose,barsoum2017hp}. Chao \etal~\cite{chao2017forecasting} proposed a 3D Pose Forecasting Network on static images; Barsoum \etal~\cite{barsoum2017hp} took a probabilistic approach for pose prediction using Wasserstein GAN \cite{arjovsky2017wasserstein}; Walker \etal~\cite{walker2017pose} proposed a variational autoencoder solution; Fragkiadaki \etal~\cite{fragkiadaki2015recurrent} proposed two architectures denoted by LSTM-3LR (3 layers of LSTM cells) and ERD (Encoder-Recurrent-Decoder); Yan \etal~\cite{yan2018mt} and Zhao \etal~\cite{zhou2018auto} proposed methods for longer time prediction; Martinez \etal~\cite{martinez2017human} used a carefully tailored RNN to learn human motion prediction; and Chiu \etal~\cite{Chiu18} proposed a multi-layer hierarchical RNN architecture (denoted by TP-RNN) to capture human dynamics. In contrast, instead of training a fully supervised model, in this work we introduce the unsupervised GAIL framework into our training process to enhance the generalizability of our learned prediction policy.

%we use the ground truth pose data as expert behavior and train a model to imitate movements for future pose prediction. 

\noindent\textbf{Reinforcement Learning for Prediction:} Methods of reinforcement learning \cite{Ng00,Lillicrap16,Sutton17,Mnih13} and imitation learning \cite{Ross11,Ho16} have been used for predicting future information in videos in different forms. For instance, DeepMimic \cite{peng2018deepmimic} proposed a physics-based method for policy generation that is capable of tracking and motion capture. Other works such as R2P2 \cite{rhinehart2018r2p2} and Tai \etal~\cite{tai2018socially} used generative models with Wasserstein reformulation to perform navigation and forward path planning. Differently, in this work we reformulate sequential pose prediction into a reinforcement learning problem and train a prediction policy through imitation learning approaches.

%InfoGAIL \cite{li2017infogail}, Causal InfoGAN \cite{kurutach2018learning}, and Procedure Planning \cite{chang2019procedure} propose perspectives of using policy generative models for predicting based on visual data. 

%\noindent\textbf{Imitation Learning:} BC, GAIL and other variants. Add papers that used IL for human motion dynamics ... using GAIL for sequential modeling 

\section{Pose Prediction as Reinforcement Learning}
\label{sec:sec3}

In this section, we introduce how to transform human pose prediction into a reinforcement learning problem \cite{Sutton17}. The core task of human pose prediction is the following: given a sequence of past pose observations $\{x_1, x_2, \ldots, x_t\}$ of length $t$ from a human subject, predict the future pose sequence $\{x_{t+1}, x_{t+2}, \ldots, x_{t+l}\}$ up to length $l$. Most previous work adopts an end-to-end approach to train either a single recurrent neural network or a pair of encoder-decoder RNNs to output the whole length-$l$ prediction sequence $\{x_{t+1}, x_{t+2}, \ldots, x_{t+l}\}$ after reading in the historical sequence $\{x_1, x_2, \ldots, x_t\}$ \cite{martinez2017human,Chiu18,jain2016structural}. In contrast, under our new reinforcement learning formulation, we evenly break the whole prediction sequence $\{x_{t+1}, x_{t+2}, \ldots, x_{t+l}\}$ into multiple steps. In each step, we only make predictions over a small window of future poses conditioned on all observed and predicted pose information so far. This transformation turns our pose prediction task into a sequential decision-making problem, where reinforcement learning algorithms can naturally come into play.

More formally, suppose we divide the whole prediction sequence into $K$ steps, then each step would correspond to a window of $m = \frac{l}{K}$ pose vectors (without loss of generality, assume that $l$ is divisible by $K$). We now define a \textbf{Markov Decision Process} (MDP) \cite{Sutton17} to model the generation of the pose prediction sequence. At each step $i$, where $i \in \{1,2,\ldots,K\}$, the state of the MDP is defined as the list of all previous pose vectors: $s_i = \{x_1, x_2, \ldots, x_{t+(i-1)\times m}\}$ and the action of the MDP is defined as the next length-$m$ window of pose vectors that the learning agent needs to predict: $a_i = \{x_{t+(i-1)\times m + 1}, \ldots, x_{t+i\times m} \}$. The transition dynamic of this MDP is deterministic, which means that at each step $i$, taking an action $a_i = \{x_{t+(i-1)\times m + 1}, \ldots, x_{t+i\times m} \}$ at state $s_i = \{x_1, x_2, \ldots, x_{t+(i-1)\times m}\}$ would deterministically transition the MDP into the new state $s_{i+1} = \{x_1, x_2, \ldots, x_{t+i\times m}\}$ at step $i+1$. The new state $s_{i+1}$ is formed by appending the action $a_i$ to the end of the current state $s_i$. This MDP process of \textbf{progressive prediction} is illustrated in Figure \ref{fig:fig2}. 

\noindent\textbf{Motivation:} There are three key motivations behind this reinforcement learning formulation of the human pose prediction problem. First, through breaking the long prediction sequence into smaller pieces of pose windows, at each step the prediction agent only needs to focus on learning to predict a much shorter period of time into the future, which greatly reduces the difficulty of prediction for the agent.
Second, the strong sequential correlation across different actions in a MDP guarantees that our agent's prediction policy will not only focus on short-term prediction accuracy but also takes long-term prediction performance into consideration. Third, only through this formulation can we apply the unsupervised GAIL approach to enhance the generalizability of our pose prediction policy over unseen domains. 

\section{Imitation Learning for Predicting Human Pose Sequences}
Now that we have formulated human pose prediction into a reinforcement learning problem, in this section, we develop an imitation learning method to train the prediction policy under this new reinforcement learning formulation.

\subsection{Deterministic Policy}

Under our formulation, the policy of our reinforcement learning agent would be a mapping from the space of all possible states $\mathcal{S}$ to the space of all possible actions $\mathcal{A}$. This policy can be either deterministic or stochastic. Under our current setting of human pose prediction, stochastic policies \cite{Sutton17} are not suitable because the dimensionality of the action space $\cal{A}$ is very high. So we decide to use deterministic policies in our imitation learning algorithms for human pose prediction. Formally, the policy of our prediction agent is a function $a = \pi_{\theta}(s)$ that maps each possible state $s$ in $\mathcal{S}$ to a single action $a$ in $\mathcal{A}$, where $\theta$ denotes the parameters of this policy mapping function.

\subsection{Behavioral Cloning}

A classical approach to imitation learning is behavioral cloning (BC) \cite{Bain95, Ross11}, where a policy is trained using supervised learning methods to minimize the distance between its generated actions and the expert's actions under the state distribution encountered by the expert. Let the expert policy function be $\pi_E$ and the state distribution encountered by the expert be $d_{\pi_E}$, then a behavioral cloning algorithm tries to find an optimal policy $\pi_{\theta^*}$ such that:
\begin{align}
\pi_{\theta^*} = \underset{\pi_{\theta} \in \Pi}{\arg\min} \,\, \mathbb{E}_{s \sim d_{\pi_E}} \big[l\big(\pi_{\theta}(s), \pi_E(s)\big)\big],
\end{align}
where $l$ is a loss function that measures the distance between two action vectors in the action space $\mathcal{A}$.

BC enjoys the advantage of being highly sample-efficient and fast to compute \cite{sammut2010behavioral}, since it aims to directly match the learning agent's policy with the expert's policy at every state that appears in the repository of expert-demonstrated trajectories. However, there are some caveats associated with BC. The first problem is the generalization issue --- the agent's policy may be overfitted to the expert's demonstrations in the areas of the state space that the expert traversed, and thus may not generalize well to areas outside the expert's experience. Second, behavioral cloning optimizes the agent's policy at each individual state separately and ignores the sequential correlation across different steps in a trajectory, which tends to make the agent's policy shortsighted. To tackle these problems, we introduce generative adversarial training into our algorithm and use an adapted version of GAIL to further optimize our agent's policy.  

\subsection{Generative Adversarial Imitation Learning}

Generative Adversarial Imitation Learning (GAIL) is a popular model-free imitation learning framework that was recently proposed by Ho and Ermon \cite{Ho16} and is inspired by the success of Generative Adversarial Networks (GAN) \cite{Goodfellow14}. At its core, GAIL takes an inverse reinforcement learning \cite{Ng00} approach to the problem of imitation learning, where it aims to first recover an estimate of the reward signals underlying the MDP from the expert's demonstration and then learns to optimize the agent's policy using the rewards signals that it recovered. Directly estimating reward functions from expert demonstration in high-dimensional and complex state-action space is intractable, so GAIL turns the inverse reinforcement learning problem into its equivalent dual problem of occupancy measure matching, where the agent seeks to match the distribution of state-action pairs (called occupancy measure) generated by its own policy to the distribution generated by the expert's policy \cite{Ho16}. In its original formulation, GAIL employs a generative adversarial training process to iteratively minimize the Jensen-Shannon divergence between the two distributions, and its learning objective is:
\vspace{-10pt}
\begin{align}
\underset{\pi_{\theta}}{\min} \,\,& \underset{D_w \, \in \, \, \mathcal{S} \times \mathcal{A} \rightarrow (0, 1)}{\max} \nonumber
\Big(\mathbb{E_{\pi_{\theta}}}[\log(D_w(s,a))] \\ & + \mathbb{E}_{\pi_{E}}[\log(1 - D_w(s,a))] - \lambda H(\pi_{\theta})\Big),
\label{eq:gail}\end{align}

where $D_w$ can be interpreted as a discriminative classifier trying to distinguish between state-action pairs generated from the agent's policy and state-action pairs generated from the expert's policy, and $H(\pi_{\theta})$ denotes the discounted causal entropy of the policy $\pi_{\theta}$. Then in analogy to the training of GANs, during actual implementation, a GAIL algorithm will alternate between a $\mathbf{D}$ step that updates the discriminator parameter $w$ to maximize $\mathbb{E_{\pi_{\theta}}}[\log(D_w(s,a))] + \mathbb{E}_{\pi_{E}}[\log(1 - D_w(s,a))]$, and a $\mathbf{G}$ step that updates the parameter $\theta$ of the policy generator $\pi_{\theta}$ using policy gradient methods such as Trust Region Policy Optimization (TRPO) \cite{Schulman15} or Proximal Policy Optimization (PPO) \cite{Schulman17} in order to minimize $\mathbb{E_{\pi_{\theta}}}[\log(D_w(s,a))] + \lambda H(\pi_{\theta})$, until the system reaches a saddle point (see Figure \ref{fig:1}). Previous work shows that GAIL often tends to generalize better than BC and achieves superior performance on complex tasks.

However, in our human pose prediction scenario, there are two major difficulties that hinder the effective application of the original GAIL algorithm. First, the human motion dynamics is highly complex, and thus our randomly initialized policy generator will be vastly different from the expert's policy at the beginning of GAIL training. This means that at the early stage of GAIL training, it is very likely that the policy generator manifold would have no non-negligible intersection with the expert manifold at all. This issue will let the Jensen-Shannon divergence to saturate quickly and cause the severe problem of vanishing gradients for the discriminator \cite{arjovsky2017wasserstein}. Second, the dimensionality of our action space $\mathcal{A}$ is very high, which makes the sampling-based policy gradient reinforcement learning algorithms like TRPO and PPO no longer applicable in practice. Therefore, we adapt GAIL to a specific new form to fit the needs of our human pose prediction task, which we call WGAIL-div.

\subsection{Adapted WGAIL-div} 
The problem of non-overlapping manifolds and vanishing gradients for the discriminator has long been studied for GANs \cite{arjovsky2017wasserstein}, and several new GAN training frameworks based on Wasserstein distances between distributions has been proposed to overcome this problem. Some of the notable ones are Wasserstein GAN (WGAN) \cite{arjovsky2017wasserstein}, Wasserstein GAN with gradient penalty (WGAN-gp) \cite{Gulrajani17}, and Wasserstein-divergence GAN (WGAN-div) \cite{Wu18}. This series of Wasserstein-based GANs changes the discriminator of GANs from a classifier that tries to distinguish between real and fake samples and outputs probability values between 0 and 1 into a critic function that directly outputs real numbers in $(-\infty, +\infty)$, and uses different methods to enforce the Lipschitz constraint on the critic function. In this work, we borrow the formulation of WGAN-div \cite{Wu18} into GAIL to construct an improved version of GAIL named \textbf{WGAIL-div}. More specifically, the objective function of our WGAIL-div algorithm is:
\begin{align}
\underset{\pi_{\theta}}{\min} \,\,& \underset{D_w \, \in \, \, \mathcal{S} \times \mathcal{A} \rightarrow \mathbb{R}}{\max}  \Big( \mathbb{E_{\pi_{\theta}}}[D_w(s,a)] - \mathbb{E}_{\pi_{E}}[D_w(s,a)] \nonumber \\
& - k \underset{(\hat{s},\hat{a}) \sim \mathbb{P}_u}{\mathbb{E}}[\|\nabla_{(\hat{s},\hat{a})} D_w((\hat{s},\hat{a}))\|^{p}] \Big),
\end{align}
where $\mathbb{P}_u$ is the distribution obtained by sampling uniformly from the straight lines connecting points of state-action pairs generated by the policy generator and points of state-action pairs in expert demonstration in the state-action space $\mathcal{S} \times \mathcal{A}$, and $k$ and $p$ are hyperparameters that adjust the level of Lipschitz-constraint regularization. Note that in this WGAIL-div objective function we no longer have the causal entropy term $H(\pi_{\theta})$ that appears in \eqref{eq:gail}, since we use deterministic policies that have constant causal entropies.

\begin{figure}
\centerline{\includegraphics[width=0.9\linewidth]{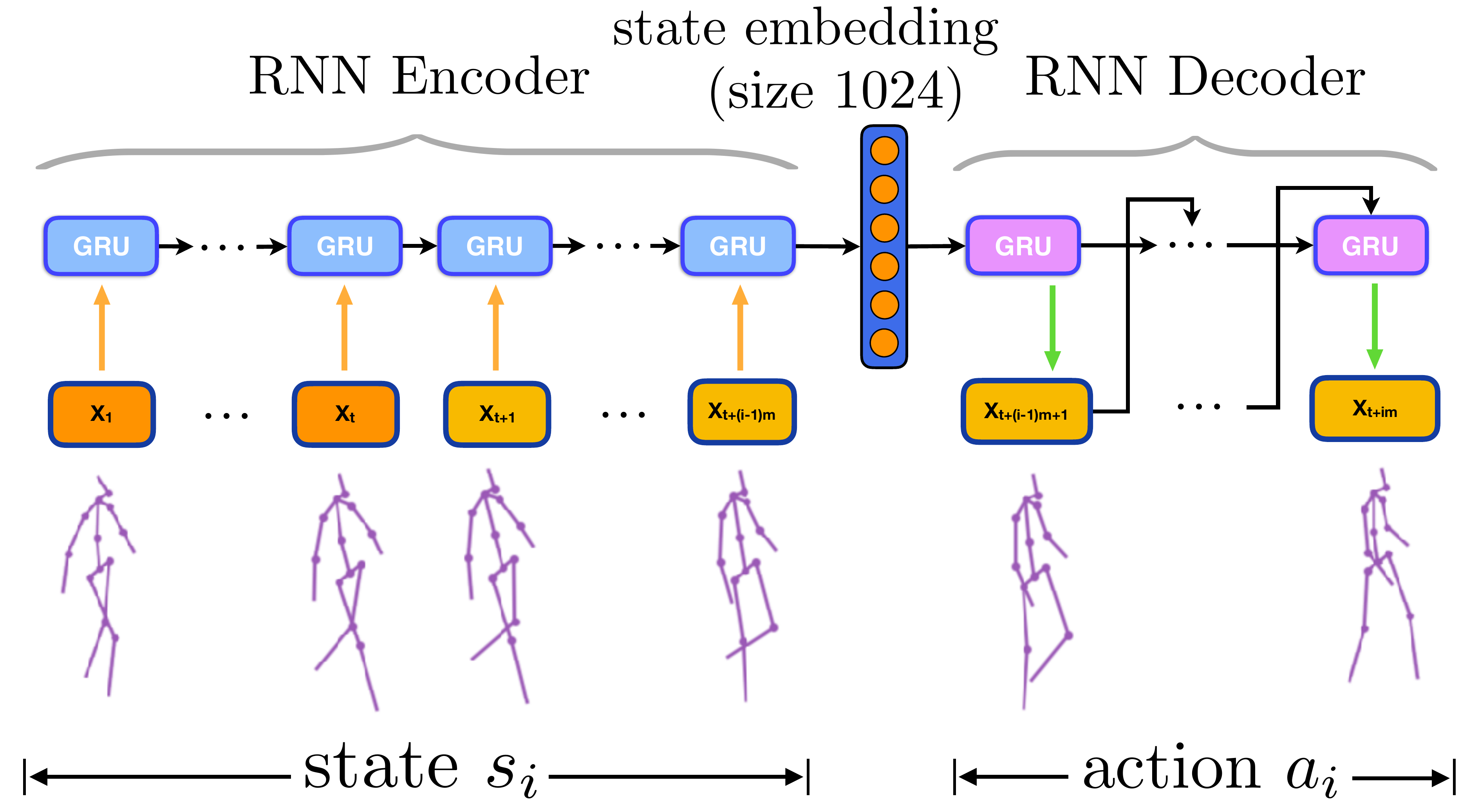}}
\vspace{-5pt}
\caption{Seq2Seq architecture of our policy generator.}
\vspace{-10pt} \label{fig:fig3}
\end{figure}

In our implementation of WGAIL-div, we alternate between a $\mathbf{D}$ step that updates the discriminator parameter $w$ to maximize $\mathbb{E_{\pi_{\theta}}}[D_w(s,a)] - \mathbb{E}_{\pi_{E}}[D_w(s,a)] - k \underset{(\hat{s},\hat{a}) \sim \mathbb{P}_u}{\mathbb{E}}[\|\nabla_{(\hat{s},\hat{a})} D_w((\hat{s},\hat{a}))\|^{p}]$, and a $\mathbf{G}$ step that updates the parameter $\theta$ of the policy generator $\pi_{\theta}$ to minimize $\mathbb{E_{\pi_{\theta}}}[D_w(s,a)]$. As mentioned in the second problem facing GAIL above, for the $\mathbf{G}$ step, the classic sampling-based approach taken by the original GAIL of using TRPO or PPO to optimize a parametrized stochastic policy fails in the high-dimensional action space $\mathcal{A}$. Therefore, inspired by \cite{Silver14} and \cite{Lillicrap16}, we use deterministic policy and substitute the TRPO step in $\mathbf{G}$ with a deterministic policy gradient step that directly use the gradient information from both the critic function $D_w$ and the policy generating function $\pi_{\theta}$ to update $\theta$ in order to minimize the expected accumulated critic loss value incurred by the agent as it makes predictions according to its policy $\pi_{\theta}$. More specifically, suppose that during the $\mathbf{G}$ step we execute our prediction agent's current policy $\pi_{\theta}$ to generate a collection of $N$ samples of state-action pairs $\{(s_i, a_i)\}_{i = 1}^{N}$, then our current estimate of the gradient over the policy parameter $\theta$ would be:
\begin{align}
& \quad \,\,\, \nabla_{\theta} \,\,\mathbb{E}_{\pi_\theta} [D_w(s, a)] \nonumber \\
& \approx \frac{1}{N} \sum_{i = 1}^{N} \nabla_a D_w(s,a)|_{s = s_i, a = \pi_{\theta}(s_i)} \nabla_{\theta} \pi_{\theta}(s)|_{s = s_i}
\end{align}

\begin{figure}
\centerline{\includegraphics[width=0.9\linewidth]{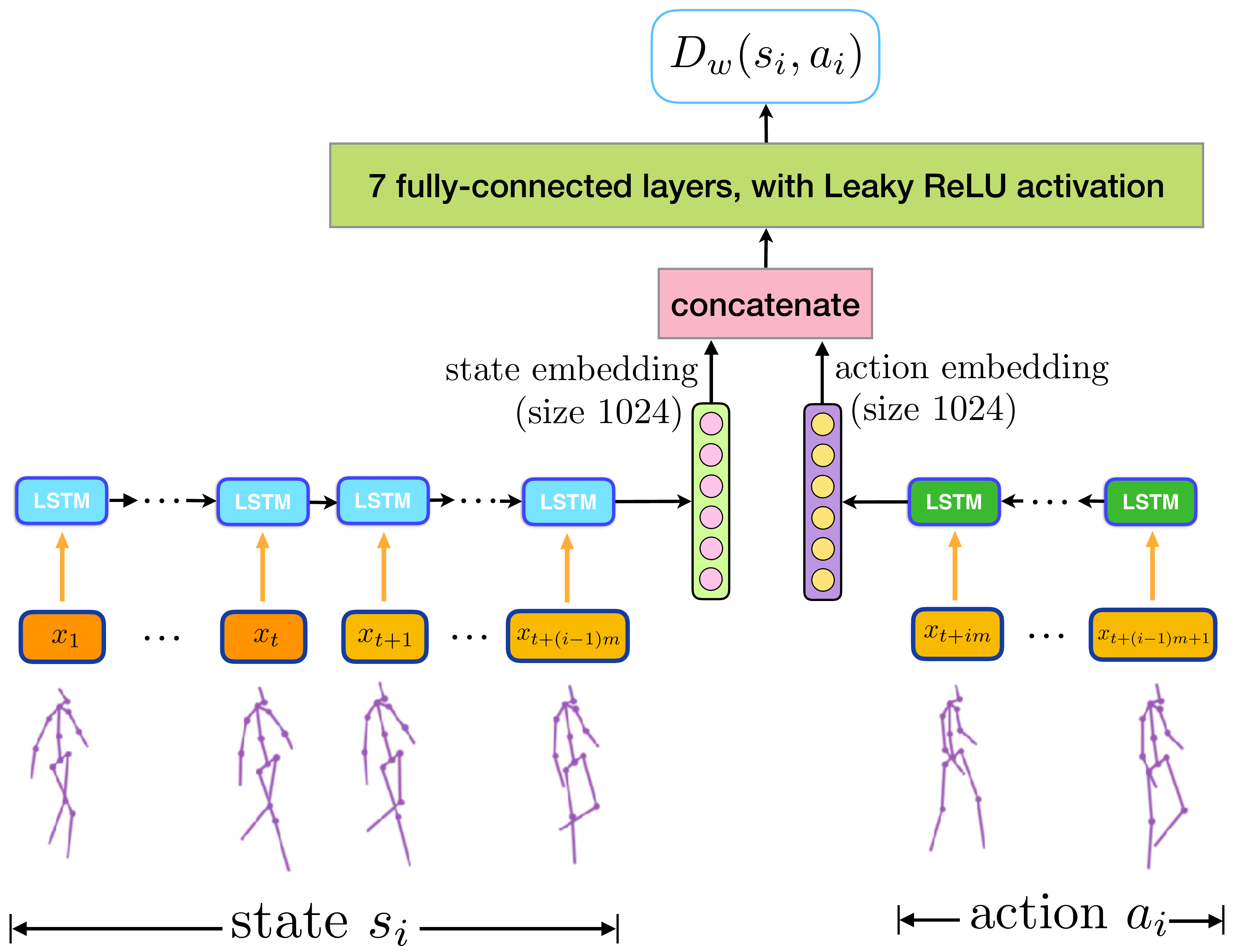}}
\vspace{-5pt}
\caption{Architecture of our critic network in WGAIL-div.}
\vspace{-10pt} \label{fig:fig4}
\end{figure}

\subsection{Model Architectures}
%In this section we describe the model architectures that we design to build our new imitation learning system for human pose prediction. 
Our imitation learning system has two components: a policy generator network $\pi_{\theta}$ and a critic network $D_w$.

\noindent\textbf{Policy Generator Network:}
The policy generator network $\pi_{\theta}$ is at the core of our imitation learning system and is responsible for predicting future human poses by sequentially generating actions (small windows of consecutive pose vectors) from states (long sequences of previous pose observations and predictions). It is used in both behavioral cloning and WGAIL-div. At each prediction step $i$, $\pi_{\theta}$ needs to read in a length-$[t + (i-1)\times m]$ sequence of state $s_i = \{x_1, x_2, \ldots, x_{t + (i-1)\times m}\}$ and then outputs a length-$m$ sequence as the action $a_i = \{x_{t + (i-1)\times m + 1}, \ldots, x_{t+i\times m} \}$. To achieve this functionality, we use the sequence-to-sequence (seq2seq) architecture \cite{Sutskever14} for our policy generator network. We construct a recurrent neural network (RNN) with a Gated Recurrent Unit (GRU) \cite{Cho14} cell of size 1024 as our encoder to read in the state sequence $s_i$, and construct another RNN with a different GRU cell of size 1024 as our decoder to generate the action sequence $a_i$. Similar to \cite{martinez2017human}, we also add an extra layer of linear spatial decoder on top of each GRU to map its 1024-dimensional output vector down to a 54-dimensional vector (54 is the length of each pose vector in our dataset, see Section \ref{sec:experiments} for details). The model architecture of $\pi_{\theta}$ is shown in Figure \ref{fig:fig3}.

\noindent\textbf{Critic Network:}
In our WGAIL-div algorithm, besides $\pi_{\theta}$, we also need to construct a critic network $D_w$ that assigns critic values to all state-action pairs $(s_i, a_i)$ in $\mathcal{S}\times \mathcal{A}$. For this critic network, we build a RNN with a Long Short-Term Memory (LSTM) \cite{Schmidhuber97} cell of size 1024 to encode the state sequence $s_i = \{x_1, x_2, \ldots, x_{t + (i-1)\times m}\}$ into a 1024-dimensional vector of state embedding $h(s_i)$, and build another RNN with a different LSTM cell of size 1024 to encode the action sequence $a_i = \{x_{t + (i-1)\times m + 1}, \ldots, x_{t+i\times m} \}$ into a 1024-dimensional vector of action embedding $u(a_i)$. We then concatenate $h(s_i)$ and $u(a_i)$ together into a 2048-dimensional vector embedding, which we feed into a 7-layer fully-connected neural network with the output size of each layer equals to 512, 256, 128, 64, 32, 16, 1, respectively and the activation functions being Leaky ReLU. This 7-layer fully-connected neural network outputs the final critic value for the state-action pair that $D_w$ reads in. The model architecture of $D_w$ is depicted in Figure \ref{fig:fig4}.

\subsection{Learning Algorithm}
Our proposed imitation learning algorithm for human pose prediction utilizes two methods: BC and WGAIL-div. They both have their respective advantages and disadvantages, and their advantages and disadvantages make them highly complementary to each other. BC has fast training speed and high sample efficiency, but may suffer from bad generalization and error compounding. In contrast, WGAIL-div often produces policies that have better generalization property and pays more attention to the sequential relationship across different time steps, but can be slow and difficult to train during the beginning iterations when the freshly initialized agent policy is very far away from the expert policy. Then under our current setting of using imitation learning to predict human motion dynamics, these characteristics indicate that BC tends to perform better over short-term predictions while WGAIL-div tends to be superior over long-term predictions. Therefore, in order to have the best of both worlds and to let our learning algorithm excel at both short-term and long-term predictions, we first train our policy generator network $\pi_{\theta}$ using BC, and then use the trained parameters to initialize $\pi_{\theta}$ in the WGAIL-div procedure. On one hand, the initialization obtained from BC would help $\pi_{\theta}$ to quickly move to regions that are close to the expert policy, which greatly stabilizes and expedites the training of WGAIL-div. On the other hand, WGAIL-div iterations can be viewed as further optimizing over the shortsighted policy generated from BC to make it generalize better to unfamiliar regions of the state space. As a result, our learning algorithm is consisted of two stages:

\begin{algorithm}[t]
  \caption{\,WGAIL-div for Human Pose Prediction} \label{alg:alg1}
  {\small
  \begin{algorithmic}[1]
    \STATE Initialize the policy generator network $\pi_{\theta}$ using the parameter $\theta$ trained by BC (Algorithm 2 in Appendix A.1)
    \STATE Randomly initialize the parameter $w$ of the critic network $D_w$
    \FOR{iteration = 1, 2, \ldots, $T$}
    \Statex \quad\underline{\textbf{D Step:}}
      \STATE Randomly sample a batch of $N_D$ length-$(t+l)$ trajectories
      \Statex \quad of human pose vectors from the training dataset $\mathcal{E}$
      \FOR{$j = 1, 2, \ldots, N_D$}
        \STATE Take the $j$-th sampled trajectory $\{x_{j,1}, x_{j,2}, \ldots,x_{j, t+l}\}$
        \FOR{$i = 1, 2, \ldots, K$}
          \STATE $s_{j, i} = \{x_{j,1}, \, \ldots\, , x_{j, t+(i-1)m}\}$,
          \STATE $a_{j, i} = \{x_{j, t+(i-1)m+1}, \,\ldots \,, x_{j, t+im}\}$
          \STATE \textbf{if} $i = 1$, \textbf{then} $\tilde{s}_{j,i} = s_{j,i}$, 
          \Statex \quad\quad\quad \textbf{else} $\tilde{s}_{j,i} = \text{concatenate}[\tilde{s}_{j,i-1}, \tilde{a}_{j,i-1}]$
          \STATE $\tilde{a}_{j,i} = \pi_\theta(\tilde{s}_{j,i})$
          \STATE Sample $\alpha$ from the uniform distribution: $\alpha \sim U[0, 1]$
          \STATE $\hat{s}_{j,i} = \alpha \, s_{j,i} + (1-\alpha) \, \tilde{s}_{j,i}$, 
          \STATE $\hat{a}_{j,i} = \alpha \, a_{j,i} + (1-\alpha) \, \tilde{a}_{j,i}$
        \ENDFOR
      \ENDFOR
      \STATE Take an Adam step on $w$ to maximize the objective:
      \STATEx \quad $w$ $\leftarrow$ Adam \Big($\nabla_{w} \, \big[ \frac{1}{N_D\times K} \,\sum_{j,i}D_w(\tilde{s}_{j,i}, \tilde{a}_{j,i})$ 
      \Statex \quad\quad\quad $- D_w(s_{j,i}, a_{j,i}) - k\, \|\nabla_{(\hat{s_{j,i}},\hat{a_{j,i}})} D_w((\hat{s_{j,i}},\hat{a_{j,i}}))\|^{p} \big] \Big)$
      \STATEx \quad \underline{\textbf{G Step:}}
      \STATE Randomly sample a batch of $N_G$ length-$(t+l)$ trajectories 
      \Statex \quad of human pose vectors from the training dataset $\mathcal{E}$
      \STATE Generate the agent's and the expert's state-action pairs \Statex \quad $\{(\tilde{s}_{j,i}, \tilde{a}_{j,i})\}$ and $\{(s_{j,i}, a_{j,i})\}$ in the same way as in \textbf{D}.
      \STATE Take an Adam step on $\theta$ to minimize the objective function:
      \STATEx \quad $\theta$ $\leftarrow$ Adam $ \Big(\frac{1}{N_G\times K} \, \sum_{j, i} \, \nabla_{a} D_w(s,a)|_{s = \tilde{s}_{j,i}, a = \pi_{\theta}(\tilde{s}_{j,i})}$ 
      \STATEx $\quad\quad\quad \textbf{}\nabla_{\theta} \pi_{\theta}(s)|_{s =  \tilde{s}_{j,i}} \Big) $
    \ENDFOR
  \end{algorithmic}
  }
\end{algorithm}

\noindent\textbf{Stage 1: Behavioral Cloning.} 
In the BC stage, we randomly sample expert trajectories from the training dataset $\mathcal{E}$, and update the parameter $\theta$ of the policy generator network $\pi_\theta$ in order to match its generated actions with the expert's actions over the states that appear in the sampled trajectories. Here, we use $\ell_1$-norm to measure the distance between two action vectors. Refer to Algorithm 2 in Appendix A.1 for the detailed steps of our BC algorithm.

\noindent\textbf{Stage 2: WGAIL-div.}
In the WGAIL-div stage, our algorithm alternates between a $\mathbf{D}$ step that updates the critic function and a $\mathbf{G}$ step that optimizes the policy generator through deterministic policy gradients. The parameter $w$ of the critic function is initialized randomly and the parameter $\theta$ of the policy generator is initialized using the trained parameter from Stage 1. Our WGAIL-div algorithm for human pose prediction is listed in Algorithm \ref{alg:alg1}.

\begin{figure}
\centerline{\includegraphics[width=\columnwidth]{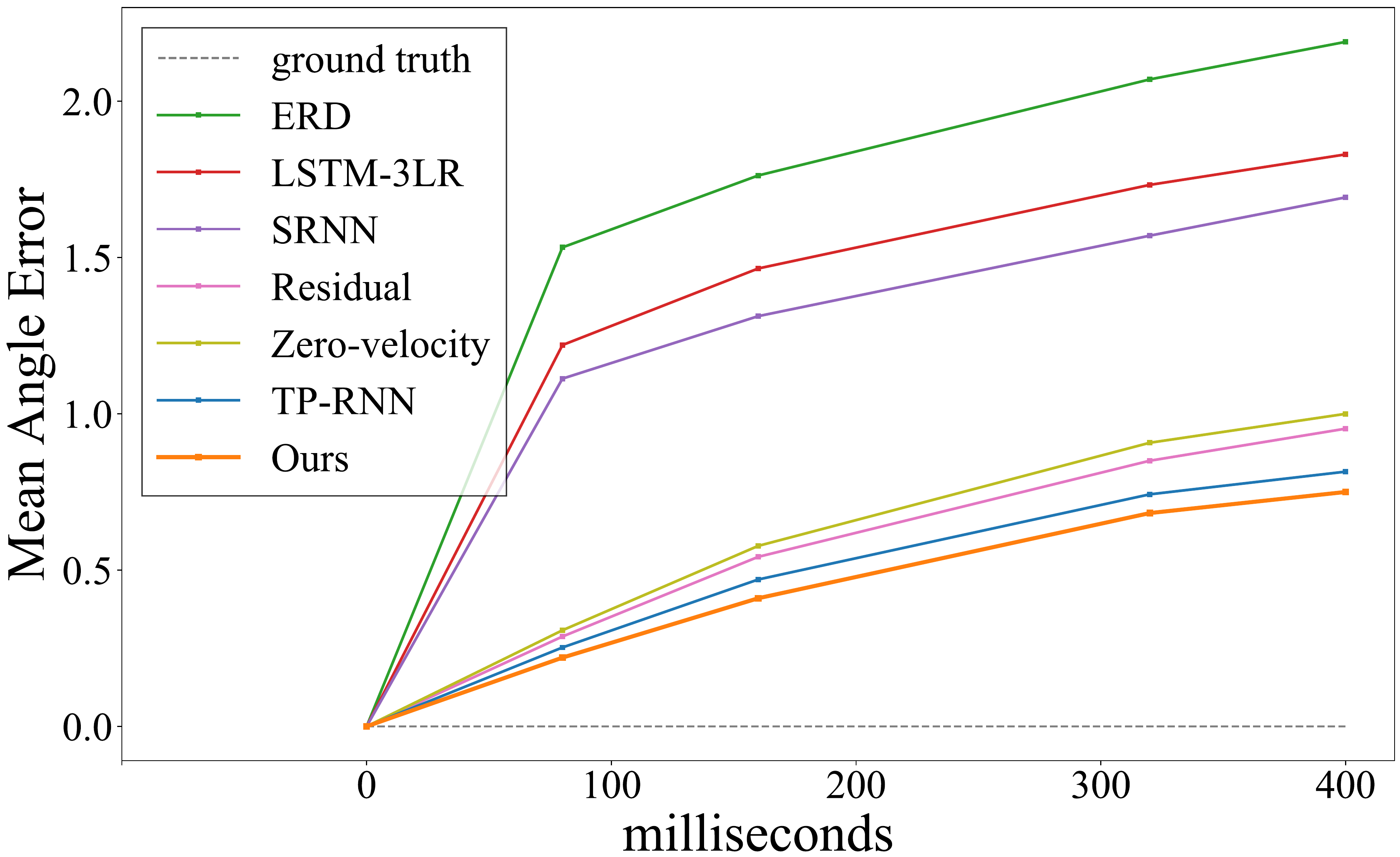}}
\caption{Plot of average mean angle error for different pose prediction models over \textit{walking}, \textit{eating}, \textit{smoking} and \textit{discussion} in the short-term prediction experiment.} \label{fig:fig5}
\vspace{-5pt}
\end{figure}

\begin{figure}
\centerline{\includegraphics[width=\columnwidth]{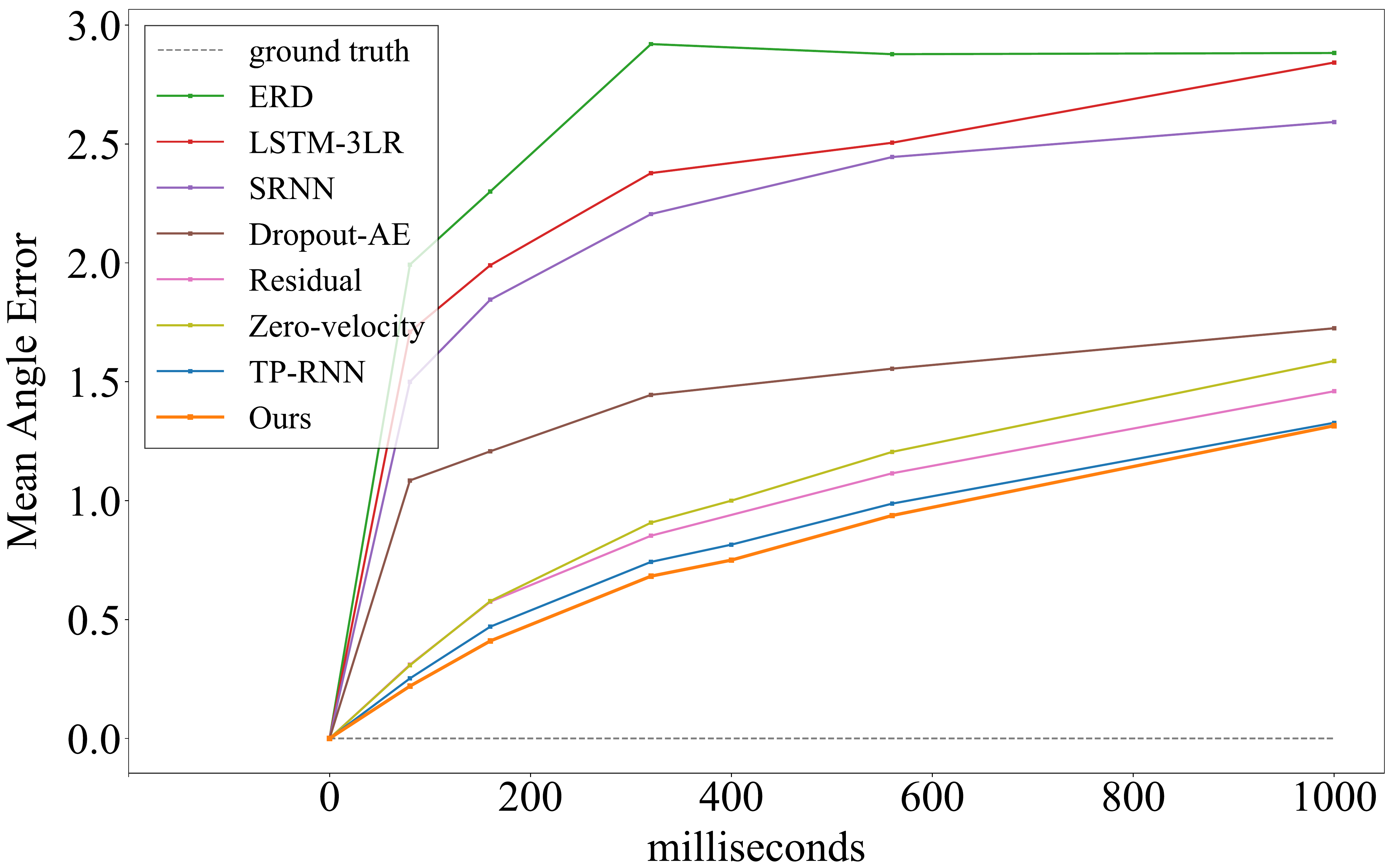}}
\caption{Plot of average mean angle error for different pose prediction models over \textit{walking}, \textit{eating}, \textit{smoking} and \textit{discussion} in the long-term prediction experiment.} \label{fig:fig6}
\vspace{-10pt}
\end{figure}

\begin{table*}[t!]
\setlength{\tabcolsep}{6.8pt}
\caption{Mean Angle Error for the activities \textit{walking}, \textit{eating}, \textit{smoking}, and \textit{discussion} in the short-term prediction experiment. The best result in each column is highlighted with boldface.}\label{tab:res1}
\vspace{-20pt}
\begin{center}
{\small
\scalebox{0.9}{
\begin{tabular}{ l|cccc|cccc|cccc|cccc  }
 \hline
 & \multicolumn{4}{c|}{Walking} & \multicolumn{4}{c|}{Eating} & \multicolumn{4}{c|}{Smoking} & \multicolumn{4}{c}{Discussion} \\
 \hline
 milliseconds               &   80 &  160 &  320 &  400 &   80 &  160 &  320 &  400 &   80 &  160 &  320 &  400 &   80 &  160 &  320 &  400 \\
 \hline
 ERD    \cite{fragkiadaki2015recurrent}       & 0.93 & 1.18 & 1.59 & 1.78 & 1.27 & 1.45 & 1.66 & 1.80 & 1.66 & 1.95 & 2.35 & 2.42 & 2.27 & 2.47 & 2.68 & 2.76\\ 
 LSTM-3LR   \cite{fragkiadaki2015recurrent}     & 0.77 & 1.00 & 1.29 & 1.47 & 0.89 & 1.09 & 1.35 & 1.46 & 1.34 & 1.65 & 2.04 & 2.16 & 1.88 & 2.12 & 2.25 & 2.23\\ 
 SRNN     \cite{jain2016structural}            & 0.81 & 0.94 & 1.16 & 1.30 & 0.97 & 1.14 & 1.35 & 1.46 & 1.45 & 1.68 & 1.94 & 2.08 & 1.22 & 1.49 & 1.83 & 1.93\\
 Residual       \cite{martinez2017human}        & 0.28 & 0.49 & 0.72 & 0.81 & 0.23 & 0.39 & 0.62 & 0.76 & 0.33 & 0.61 & 1.05 & 1.15 & 0.31 & 0.68 & 1.01 & 1.09 \\
 Zero-velocity     \cite{martinez2017human}                         & 0.39 & 0.68 & 0.99 & 1.15 & 0.27 & 0.48 & 0.73 & 0.86  & 0.26 & 0.48 & 0.97 & 0.95 & 0.31 & 0.67 & 0.94 & 1.04 \\
 TP-RNN        \cite{Chiu18}           & 0.25 & 0.41 & 0.58 & 0.65 & 0.20 & 0.33 & 0.53 & 0.67 & 0.26 & 0.47 & 0.88 & 0.90 & 0.30 & 0.66 & 0.96 & 1.04 \\
 Ours                    & \textbf{0.21} & \textbf{0.34} & \textbf{0.53} & \textbf{0.59} & \textbf{0.17} & \textbf{0.30} & \textbf{0.52} & \textbf{0.65} & \textbf{0.23} & \textbf{0.44} & \textbf{0.87} & \textbf{0.85} & \textbf{0.23} & \textbf{0.56} & \textbf{0.82} & \textbf{0.91} \\
 \hline
\end{tabular}
}
}
\end{center}
\vspace{-10pt}
\end{table*}

\begin{table*}[t!]
\caption{Mean Angle Error for the activities \textit{walking}, \textit{eating}, \textit{smoking}, and \textit{discussion} 
in the long-term prediction experiment. The best result in each column is highlighted with boldface.}\label{tab:res2}
\begin{center}
\vspace{-20pt}
\setlength{\tabcolsep}{3.8pt}
{\small
\scalebox{0.9}{
\begin{tabular}{ l|ccccc|ccccc|ccccc|ccccc }
 \hline
 & \multicolumn{5}{c|}{Walking} & \multicolumn{5}{c}{Eating} & \multicolumn{5}{c|}{Smoking} & \multicolumn{5}{c}{Discussion} \\
 \hline
 milliseconds               &   80 &  160 &  320 &  560 & 1000 &   80 &  160 &  320 &  560 & 1000    &   80 &  160 &  320 &  560 & 1000 &   80 &  160 &  320 &  560 & 1000 \\
 \hline
 ERD       \cite{fragkiadaki2015recurrent}                  & 1.30 & 1.56 & 1.84 & 2.00 & 2.38 & 1.66 & 1.93 & 2.88 & 2.36 & 2.41 & 2.34 & 2.74 & 3.73 & 3.68 & 3.82 & 2.67 & 2.97 & 3.23 & 3.47 & 2.92\\ 
 LSTM-3LR     \cite{fragkiadaki2015recurrent}              & 1.18 & 1.50 & 1.67 & 1.81 & 2.20 & 1.36 & 1.79 & 2.29 & 2.49 & 2.82  & 2.05 & 2.34 & 3.10 & 3.24 & 3.42 & 2.25 & 2.33 & 2.45 & 2.48 & 2.93 \\ 
 SRNN         \cite{jain2016structural}               & 1.08 & 1.34 & 1.60 & 1.90 & 2.13 & 1.35 & 1.71 & 2.12 & 2.28 & 2.58 & 1.90 & 2.30 & 2.90 & 3.21 & 3.23 & 1.67 & 2.03 & 2.20 & 2.39 & 2.43\\
 Dropout-AE    \cite{Ghosh17}       & 1.00 & 1.11 & 1.39 & 1.55 & 1.39 & 1.31 & 1.49 & 1.86 & 1.76 & 2.01  & 0.92 & 1.03 & 1.15 & 1.38 & 1.77 & 1.11 & 1.20 & 1.38 & 1.53 & \textbf{1.73}\\ 
 Residual   \cite{martinez2017human}     & 0.32 & 0.54 & 0.72 & 0.86 & 0.96 & 0.25 & 0.42 & 0.64 & 0.94 & 1.30  & 0.33 & 0.60 & 1.01 & 1.23 & 1.83 & 0.34 & 0.74 & 1.04 & 1.43 & 1.75 \\
 Zero-velocity      \cite{martinez2017human}          & 0.39 & 0.68 & 0.99 & 1.35 & 1.32 & 0.27 & 0.48 & 0.73 & 1.04 & 1.38 & 0.26 & 0.48 & 0.97 & 1.02 & 1.69 & 0.31 & 0.67 & 0.94 & 1.41 & 1.96\\
 TP-RNN   \cite{Chiu18}        & 0.25 & 0.41 & 0.58 & 0.74 & 0.77 & 0.20 & 0.33 & 0.53 & 0.84 & 1.14 & 0.26 & 0.48 & 0.88 & 0.98 & 1.66 & 0.30 & 0.66 & 0.98 & 1.39 & 1.74\\
 Ours & \textbf{0.21} & \textbf{0.34} & \textbf{0.53} & \textbf{0.67} & \textbf{0.69} & \textbf{0.17} & \textbf{0.30} & \textbf{0.52} & \textbf{0.79} & \textbf{1.13} & \textbf{0.23} & \textbf{0.44} & \textbf{0.86} & \textbf{0.95} & \textbf{1.63} & \textbf{0.27} & \textbf{0.56} & \textbf{0.82} & \textbf{1.34} & 1.81\\
\hline
\end{tabular}
}
}
\end{center}
\vspace{-15pt}
\end{table*}

\section{Experiments} 
\label{sec:experiments}
To evaluate the performance of our proposed imitation learning approach on challenging real human pose prediction tasks and to have a thorough comparison with previous works, we run exhaustive experiments to test our algorithm using the popular Human 3.6M dataset \cite{Ionescu14} and compare our results with previous benchmarks.

\noindent\textbf{The Human 3.6M dataset:}
The Human 3.6M dataset \cite{Ionescu14} is currently one of the largest publicly available motion capture dataset, and has been used by many previous papers on human pose prediction. This dataset includes video sequences recorded from 7 different actors performing 15 different categories of human activities, with each actor performing each activity in two different trials. The videos are recorded in 50\textit{Hz}, and for a fair comparison, we follow previous papers \cite{jain2016structural, martinez2017human} to downsample the pose sequence by 2. Each pose in this dataset is represented as an exponential map representation of 32 human joints in 3D, and after preprocessing of global translation and rotation as described in \cite{jain2016structural, martinez2017human, fragkiadaki2015recurrent, Chiu18}, we adopt the following evaluation methods for our experiments: during both training and testing, we feed 2000\textit{ms} (50 frames) of past pose vector sequence into the learning system, and the goal is to predict the next 1000\textit{ms} (25 frames) of future pose vector sequence. We measure the Euclidean distance between predicted poses and ground truth poses in angle space as the evaluation metric of prediction error, and we report the average prediction error over 6 different timescales: 80, 160, 320, 400, 560, and 1000\textit{ms}. Following previous works, we use Subjects 1, 6, 7, 8, 9, 11 as the training dataset and use Subject 5 as the testing dataset.

\noindent\textbf{Baselines:}
We compare our experimental results with the following recent state-of-the-art methods for human pose prediction on the Human 3.6M dataset: ERD \cite{fragkiadaki2015recurrent}, LSTM-3LR \cite{fragkiadaki2015recurrent}, SRNN \cite{jain2016structural}, Dropout-AutoEncoder \cite{Ghosh17}, Residual \cite{martinez2017human}, Zero-velocity \cite{martinez2017human,Chiu18} and TP-RNN \cite{Chiu18}.

\begin{figure}
\begin{subfigure}{\columnwidth}
\centerline{\includegraphics[width=\columnwidth]{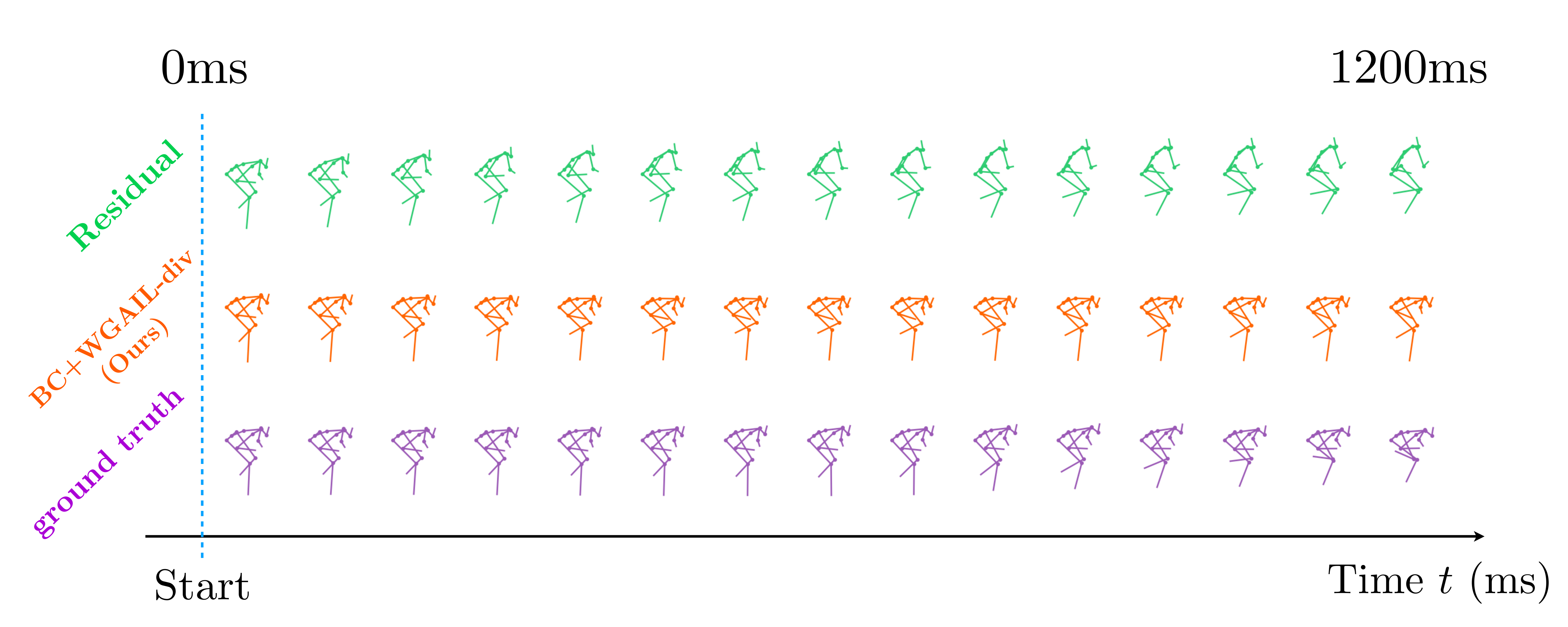}}
\caption{Taking Photo}
\end{subfigure}
\newline
\begin{subfigure}{\columnwidth}
\centerline{\includegraphics[width=\columnwidth]{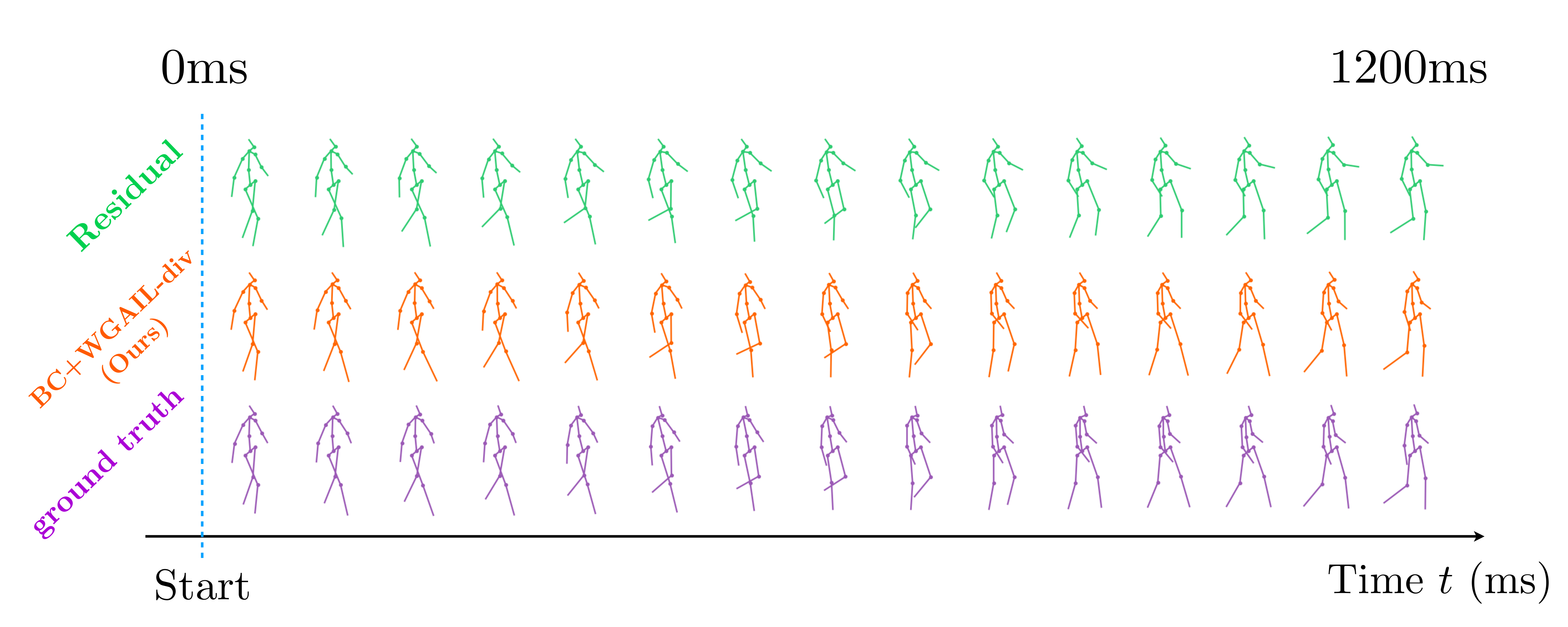}}
\caption{Walking}
\end{subfigure}
\caption{Visualization of long-term human pose prediction results for \textit{taking photo} (top) and \textit{walking} (bottom). The purple poses are ground truth, the green poses are predicted by the \textit{Residual} \cite{martinez2017human} model and the orange poses are ours.}
\label{fig:fig7}\vspace{-15pt}
\end{figure}

\begin{table*}[t!]
\caption{Mean Angle Error for the remaining 11 actions in Human 3.6M in the long-term prediction experiment. The best result in each column is highlighted with boldface.}
\label{tab:res3}
\vspace{-20pt}
\setlength{\tabcolsep}{2.5pt}
\begin{center}
{\footnotesize
\begin{tabular}{ l|cccccc|cccccc|cccccc|cccccc }
 \hline

& \multicolumn{6}{c|}{Purchases} & \multicolumn{6}{c|}{Sitting} & \multicolumn{6}{c|}{Sitting down} & \multicolumn{6}{c}{Taking photo} \\
 \hline
 millisec                &   80 &  160 &  320 & 400 &  560 & 1000 &   80 &  160 &  320 & 400 &  560 & 1000 &   80 &  160 &  320 & 400 &  560 & 1000 &   80 &  160 &  320 & 400 &  560 & 1000 \\
 \hline
Residual        & 0.60 & 0.86 & 1.24 & 1.30 & 1.58 & 2.26 & 0.44 & 0.74 & 1.19 & 1.40 & 1.57 & 2.03 & 0.51 & 0.93 & 1.44 & 1.65 & 1.94 & 2.55 & 0.33 & 0.65 & 0.97 & 1.09 & 1.19 & 1.47 \\
 TP-RNN        & 0.59 & 0.82 & 1.12 & 1.18 & 1.52 & 2.28 & 0.41 & 0.66 & 1.07 & 1.22 & 1.35 & 1.74 & 0.41 & 0.79 & 1.13 & 1.27 & 1.47 & 1.93 & 0.26 & 0.51 & 0.80 & 0.95 & 1.08 & 1.35 \\
 Ours          & \textbf{0.54} & \textbf{0.78} & \textbf{1.07} & \textbf{1.14} & \textbf{1.46} & \textbf{2.23} & \textbf{0.29} & \textbf{0.48} & \textbf{0.87} & \textbf{1.04} & \textbf{1.21} & \textbf{1.58} & \textbf{0.34} & \textbf{0.68} & \textbf{1.01} & \textbf{1.14} & \textbf{1.34} & \textbf{1.78} & \textbf{0.17} & \textbf{0.37} & \textbf{0.60} & \textbf{0.72} & \textbf{0.84} & \textbf{1.06} \\
 \hline  \hline

 & \multicolumn{6}{c|}{Directions} & \multicolumn{6}{c|}{Greeting} & \multicolumn{6}{c|}{Talking on the phone} & \multicolumn{6}{c}{Posing} \\
 \hline
 millisec                &   80 &  160 &  320 & 400 & 560 & 1000 &   80 &  160 &  320 & 400 & 560 & 1000 &   80 &  160 &  320 & 400 & 560 & 1000 &   80 &  160 &  320 & 400 & 560 & 1000 \\
 \hline
Residual       & 0.44 & 0.69 & 0.83 & 0.94 & 1.03 & 1.49 & 0.53 & 0.88 & 1.29 & 1.45 & 1.72 & 1.89 & 0.61 & 1.12 & 1.57 & 1.74 & 1.59 & 1.92 & 0.47 & 0.87 & 1.49 & 1.76 & 1.96 & \textbf{2.35} \\
 TP-RNN        & 0.38 & 0.59 & \textbf{0.75} & \textbf{0.83} & \textbf{0.95} & \textbf{1.38} & 0.51 & 0.86 & 1.27 & 1.44 & 1.72 & 1.81 & 0.57 & 1.08 & 1.44 & 1.59 & \textbf{1.47} & 1.68 & 0.42 & 0.76 & 1.29 & 1.54 & \textbf{1.75} & 2.47 \\
 Ours      & \textbf{0.27} & \textbf{0.46} & 0.81 & 0.89 & \textbf{0.95} & 1.41 & \textbf{0.43} & \textbf{0.75} & \textbf{1.17} & \textbf{1.33} & \textbf{1.62} & \textbf{1.72} & \textbf{0.54} & \textbf{1.05} & \textbf{1.40} & \textbf{1.56} & 1.52 & \textbf{1.67} & \textbf{0.27} & \textbf{0.55} & \textbf{1.16} & \textbf{1.41} & 1.83 & 2.69 \\
 \hline \hline
 
 & \multicolumn{6}{c|}{Waiting} & \multicolumn{6}{c|}{Walking dog} & \multicolumn{6}{c|}{Walking together} & \multicolumn{6}{c} {Average over all 15 actions} \\
 \hline
 millisec                &   80 &  160 &  320 & 400 &  560 & 1000 &   80 &  160 &  320 & 400 &  560 & 1000 &   80 &  160 &  320 & 400 &  560 & 1000 &   80 &  160 &  320 & 400 &  560 & 1000 \\
 \hline
Residual     & 0.34 & 0.65 & 1.09 & 1.28 & 1.61 & \textbf{2.27} & 0.56 & 0.95 & 1.28 & 1.39 & 1.68 & 1.92 & 0.31 & 0.61 & 0.84 & 0.89 & 1.00 & 1.43 & 0.43 & 0.75 & 1.11 & 1.24 & 1.42 & 1.83  \\
 TP-RNN           & 0.30 & 0.60 & 1.09 & 1.31 & 1.71 & 2.46 & 0.53 & 0.93 & 1.24 & 1.38 & 1.73 & 1.98  & 0.23 & 0.47 & 0.67 & 0.71 & 0.78 & 1.28 & 0.37 & 0.66 & 0.99 & 1.11 & 1.30 & 1.71 \\
 Ours        & \textbf{0.25} & \textbf{0.53} & \textbf{0.96} & \textbf{1.19} & \textbf{1.59} & 2.39 & \textbf{0.46} & \textbf{0.80} & \textbf{1.12} & \textbf{1.31} & \textbf{1.65} & \textbf{1.86}  & \textbf{0.19} & \textbf{0.41} & \textbf{0.61} & \textbf{0.64} & \textbf{0.67} & \textbf{1.15} & {\textbf{0.31}}  & {\textbf{0.57}} & {\textbf{0.90}} & {\textbf{1.02}} & {\textbf{1.23}} & {\textbf{1.65}} \\
 \hline
\end{tabular}
      }
\end{center}
\vspace{-15pt}
\end{table*}

\subsection{Results}

The performance results published by previous papers are reported on slightly different prediction timescales (some report on short-term predictions over 400ms and the others report on long-term predictions over 1000ms) and activity categories (some only report on 4 different activities). To make a thorough comparison with all these previous methods, we report the prediction results of our model in Tables \ref{tab:res1}, \ref{tab:res2} and \ref{tab:res3}, and plot the mean angle error curves in Figures \ref{fig:fig5} and \ref{fig:fig6}.

From the experimental results, we see that the prediction performance of our proposed imitation learning approach (BC+WGAIL-div) surpasses all the baseline models by significant margins on almost all 15 human activity categories across all different prediction timescales. Our prediction results set the new state-of-the-art performance for the task of predicting human motion on the Human 3.6M dataset over both short-term and long-term predictions.  

We also plot visualization of the long-term human pose prediction results in Figure \ref{fig:fig7} for two representative human activities: \textit{taking photo} and \textit{walking}. From the visualization we can see that the predictions made by our imitation learning algorithm look very similar to the ground truth and are much better and more natural than the predictions made by the baseline model \textit{Residual} \cite{martinez2017human}. See Appendix A.3 for more visualization of our prediction results.

\noindent\textbf{Training Speed:}
During our experiments, we observed that the training speed of our proposed imitation learning method is much faster than previous methods. In our experiments using 4 NVIDIA Titan GPUs, on average, our BC + WGAIL-div algorithm finishes training within 20 minutes, while the baseline models \textit{Residual} and \textit{TP-RNN} finish training in more than 8 hours. There is at least a $\times 24$ speedup on training using our proposed method.

\begin{table}[t]
 \caption{Ablation Study: Mean Angle Error of Behavioral Cloning (BC) alone, WGAIL-div alone, and WGAIL-div with Behavioral Cloning as pre-training (BC+WGAIL-div) averaged over the four activities \textit{walking}, \textit{eating}, \textit{smoking}, and \textit{discussion} 
in the human pose prediction experiment on the Human 3.6M dataset. The best result in each column is highlighted with boldface.} \label{tab:ablation}

{\small
\begin{tabular}{ l|cccccc }
 \hline

 millisec                &   80 &  160 &  320 & 400 &  560 & 1000 \\
 \hline
BC        & 0.23 & 0.43 & 0.72 & 0.78 & 0.96 & 1.36 \\
 WGAIL-div        & 0.23 & 0.44 & 0.74 & 0.80 & 0.98 & 1.33 \\
 BC+WGAIL-div         & \textbf{0.21} & \textbf{0.41} & \textbf{0.69} & \textbf{0.75} & \textbf{0.94} & \textbf{1.32} \\
 \hline
 \end{tabular}}

\end{table}

\subsection {Ablation Study}

Our proposed imitation learning algorithm is composed of two components: (1) a BC component (Algorithm 2 in Appendix A.1) as pretraining; and (2) a WGAIL-div component (Algorithm 1). In our ablation study, in order to test the importance of each component, we train our human pose prediction agent on the Human 3.6M dataset \cite{ionescu2014human3} using: (a) BC alone; (b) WGAIL-div alone; and (c) WGAIL-div with BC as pretraining. The results of our ablation study experiments are summarized in Table \ref{tab:ablation}. As we can see from Table \ref{tab:ablation}, removing either BC or WGAIL-div from our imitation learning algorithm would worsen the prediction performance. Therefore, our ablation study shows that both components of our imitation learning algorithm are necessary in order to achieve optimal performance.

\subsection{Discussion}

In this work, all the pose predictions are solely based on past pose observations. In our future work, we plan to extend our work to tackle more sophisticated scenarios of human motion prediction. First, we plan to extend our WGAIL-div framework to multi-agent WGAIL-div in order to take into account other humans in the surroundings when performing group human motion prediction. Second, we plan to introduce latent variables into WGAIL-div in order to effectively model other factors such as environments, objects, activities, and intentions that might also be involved in human motion prediction.

\section{Conclusion}
In this work, we proposed a new reinforcement learning formulation of the human pose prediction problem, and developed an imitation learning algorithm for pose prediction under this new formulation. Our experiments showed that our proposed method can effectively learn to make accurate human pose predictions over both short terms and long terms with very fast training speed. Our work also has great potential to be generalized to other tasks of sequential modeling and time-series predictions in computer vision and machine learning, which will also be the direction of our future work. 

\noindent\textbf{Acknowledgments:} We would like to thank Mindtree and Panasonic for their support.

{\small
\bibliographystyle{ieee_fullname}
\bibliography{refs}
}

\end{document}